\documentclass{article}

\usepackage[T1]{fontenc}
\usepackage[utf8]{inputenc}
\usepackage{amsmath}
\usepackage{float}
\usepackage{url}
\usepackage{graphicx}
\usepackage{multirow}
\usepackage{slashbox}
\usepackage{nicefrac}
\usepackage{authblk}
\usepackage{url}

\title{Data-driven forecasting of solar irradiance}

\author{Pierrick Bruneau}
\author{Philippe Pinheiro}
\author{Yoann Didry}

\affil{Luxembourg Institute of Science and Technology \\ 5, Avenue des Hauts-Fourneaux \\
L-4362 Esch-sur-Alzette \\ mail: \emph{firstname.lastname@list.lu}}

\date{}

\begin{document}

\maketitle

\begin{abstract}
This paper\footnote{This work has been published in French with the title \emph{Prédiction du Rayonnement Solaire par Apprentissage Automatique}, in the EGC 2018 conference, see \url{https://egc18.sciencesconf.org/}} describes a flexible approach to short term prediction of meteorological variables. In particular, we focus on the prediction of the solar irradiance one hour ahead, a task that has high practical value when optimizing solar energy resources. As \emph{Défi EGC 2018} provides us with time series data for multiple sensors (e.g. solar irradiance, temperature, hygrometry), recorded every minute for two years and 5 geographical sites from \emph{La Réunion} island, we test the value of using recently observed data as input for prediction models, as well as the performance of models across sites.
After describing our data cleaning and normalization process, we combine a variable selection step based on \emph{AutoRegressive Integrated Moving Average} (ARIMA) models, to using general purpose regression techniques such as neural networks and regression trees.
\end{abstract}

\section{Introduction}
In this paper, we focus on the problem of forecasting the future value of a time series of interest for arbitrary horizons, using multivariate time series of observed sensor values as input. Specifically, in the context of \emph{Défi EGC 2018}, we consider the prediction of future solar irradiance using its present and past values, along with other meteorological variables (e.g. air temperature) obtained from sensors at fixed time steps.
Prediction of future values of meteorological variables has important value when optimizing the usage of renewable energy sources \cite{barbounis06}.

The most classical approach to meteorological forecasting uses physical models initialized with values collected for the present time using a geographical array of sensors \cite{lynch08}. Alternatively, in this paper we consider a purely data-driven approach, that completely ignores the physics underlying the time series, and views the forecasting task as a general regression problem. In this context, the target variable is the future value for the time series of interest, and potential input variables are made from the sensor values up to the present time.
Generally, adopting such a data-driven perspective yields the following sub-problems:

\begin{itemize}
\item \emph{data pre-processing}: missing values generally bode ill for the application of machine learning methods. Careful inspection and correction of data is necessary prior to such activities.
\item \emph{variable selection}: with time series, balance has to be found between the most naive model that uses only the current sensor values as input, and the most complex, that uses the complete data history. The latter can intuitively bring improvements, but estimating many parameters yields sparsity and complexity issues.
\end{itemize}

After reviewing related work in Section \ref{sec:related}, we describe the data sets of \emph{Défi EGC 2018} in Section \ref{sec:data}\footnote{The data can be downloaded at \url{http://www.egc.asso.fr/news-details-1-40-Defi_EGC_2018_Un_defi_sous_le_soleil_de_lle_de_La_Reunion}}, as well as pre-processing steps applied to them in Section \ref{sec:cleaning}. A variable selection procedure is then presented in Section \ref{sec:selection}. The consolidated data sets resulting from this pre-processing pipeline are then fed to general purpose regression models, following the protocol described in Section \ref{sec:experiments}. Experimental results then disclosed evaluate the performance of two regression models, the value of the variable selection procedure, as well as the impact of the multiple geographical sites underlying the data.

\section{Related Work} \label{sec:related}

Let us consider a set of meteorological time series, $x$ denoting one of the series.
Series are indexed by time step $t$, so that $x_t$ is the value of a given time series at a given time. A fixed time step is assumed, i.e. the time between $x_t$ and $x_{t+1}$ is constant for all $t$.
For convenience, and consistency with the literature on time series, we define the $l$-step-ahead prediction task for time series $x$ as that of predicting $x_{t+l-1}$ with knowledge of the series up to $x_{t-1}$. The 1-step-ahead prediction is then about predicting $x_t$ with knowledge up to $x_{t-1}$. In other words, for notational convenience, in this paper the present time is assumed to be $t-1$.
Considering a set of $D$ time series $\{ x_d \}_{d \in 1 \dots D}$, among which we isolate a time series of interest with index $\delta \in 1 \dots D$, we define multivariate time series prediction as the task of forecasting $x_{\delta, t+l-1}$ with the knowledge of series up to $x_{d, t-1}$ for all $d$.

ARIMA models are the seminal approach for handling univariate time series. Optimization is performed w.r.t. 1-step-ahead forecasts. They mix an auto-regressive part (i.e. $AR(p)$ process):

\begin{equation}
x_t = \sum_{i=1}^p \phi_i x_{t-i} + \sigma_t \label{eq:arp}
\end{equation}
with a moving average part (i.e. $MA(q)$ process):

\begin{equation}
\sigma_t = \sum_{i=1}^q \theta_i \sigma_{t-i} \label{eq:maq}
\end{equation}

In Eqn. \eqref{eq:arp}, $\sigma_t$ is assumed to be i.i.d. Gaussian noise. Intuitively, the forecast in Eqn. \eqref{eq:arp} is a linear combination of $p$ previously observed values in the series (i.e. $x_{t-1}$ to $x_{t-p}$), with independent random residuals. The moving average part allows dependent residuals with the linear combination of $q$ previous residuals. Assuming some technical conditions are respected, both resulting processes are stationary. An important consequence is that their expected value and residuals are constant w.r.t. $t$. ARIMA adds an integration part to the combination of these processes, allowing its use for time series with trends.

\emph{Generalized Additive Models} (GAM) \cite{hastie00} are a drastically different approach to time series modelling, that rely on summing smooth functions. It has already been used in a context reminiscent to our work with air pollution data in \cite{dominici02}.
In this view, the optimization is performed in theory w.r.t. all predicting horizons.
However GAM works on the idea of modelling series with smooth functions, which does not help much for variable selection. Most machine learning models indeed work using a multidimensional numerical vector with fixed size as input.

Various strategies could be used to adapt neural networks to a multivariate time series prediction task. A kind of divide-and-conquer approach based on classical \emph{Multi-Layer Perceptrons} (MLP) has been used in \cite{bruneau12a}.
Whereas established methodologies exist to determine which past values of the time series are useful inputs for prediction with ARIMA models (e.g. Box-Jenkins \cite{anderson76,box15}, Hyndman and Khandakar \cite{hyndman07}, among other), no straightforward method exists for neural networks.
A Bayesian selection method has been proposed for MLP models in \cite{bruneau12b}, but this requires initial training of neural networks with deliberately large input vectors, over which selection is performed a posteriori.

\section{Proposed Approach}

\subsection{Data} \label{sec:data}

The $D$ time series defined by \emph{Défi EGC 2018} are listed with their respective units as:
\begin{itemize}
  \item $I_D$: the diffuse solar irradiance ($W.m^{-2}$)
  \item $I_G$: the global solar irradiance ($W.m^{-2}$)
  \item \emph{Patm}: the atmospheric air pressure ($hPa$)
  \item \emph{RH}: the relative humidity rate (\%)
  \item \emph{Text}: the external temperature ($^{\circ}C$)
  \item \emph{WD}: the wind direction ($^{\circ}$)
  \item \emph{WS}: the wind speed ($m.s^{-1}$)
\end{itemize}

All these series have 1-minute time steps, with all $t$'s specifying exact minute timestamps (i.e. respective \emph{seconds} fields are always 0). They are recorded over 2 full years (2014 and 2015), and for 5 geographical sites, reported in Table \ref{tab:sites}.

\begin{table}
\begin{center}
\begin{tabular}{|l|l|l|l|l|}
  \hline
  Moufia & Possession & Saint André & Saint Leu & Saint Pierre \\
  \hline
  -20.92, 55.48  & -20.93, 55.33 & -20.96, 55.62 & -21.20, 55.30 & -21.31, 55.45 \\
  \hline
\end{tabular}
\end{center}
\caption{GPS coordinates (latitude, longitude) of the sites where data has been recorded.}
\label{tab:sites}
\end{table}

To better handle the manifold of angles underlying \emph{WD} (i.e. an angle of $350^{\circ}$ is closer to $10^{\circ}$ than $120^{\circ}$), we take its cosine and sine as \emph{UnitX} and \emph{UnitY} variables, respectively.
As means to facilitate the account of the strong daily and seasonal structure of the irradiance variables, we consider the \emph{direct-to-global irradiance ratio} $k_b$ (used e.g. in \cite{kylling00}). Also, this variable has a natural interpretation (i.e. 0 when cloudy, 1 when sunny), that is relevant in photovoltaic panels applications \cite{tapakis16}.
The ratio is defined as:

\begin{equation}
  k_b = \frac{I_G-I_D}{I_G} = 1 - \frac{I_D}{I_G}
\end{equation}

Some other way to normalize irradiance values would be to account for theoretical maximal irradiance models \cite[Section 2.3]{reno12}.
However such models exist only for solar irradiance. For arbitrary meteorological variables, monthly-hourly sets may serve to estimate means and deviations from training data, then used for normalization \cite{bruneau12a}.

In this paper, we focus on hourly predictions, i.e. on predicting the solar irradiance one hour later from a given time. As the time series data has 1-minute time steps, according to the terminology introduced in Section \ref{sec:related} we focus on predicting $x_{t+59}$.

\subsection{Pre-processing} \label{sec:cleaning}

Before addressing the prediction task, we investigated the provided data files in an exploratory approach. The files were pre-processed using the \emph{zoo} \cite{zeileis05} and \emph{lubridate} \cite{grolemund11} R packages, that provide convenient facilities for handling time series.

First, we checked for misspecified timestamps, missing values, and missing timestamps. If all timestamps recorded in the files appeared to be correct (i.e. correct format, and exact minute timestamps), missing values and timeslots were found. For example, in the data for \emph{Moufia}, lines for the time slots \texttt{2014-01-22 08:33:00} to \texttt{2014-01-22 08:57:00} are absent. Lines almost always specify all the variables, with the notable exception of \emph{Saint André}, where only $I_G$ is missing between \texttt{2014-01-05 08:59:00} and \texttt{2014-01-05 14:59:00}.

\begin{figure}
\begin{center}
 \includegraphics[width=0.95\textwidth, keepaspectratio]{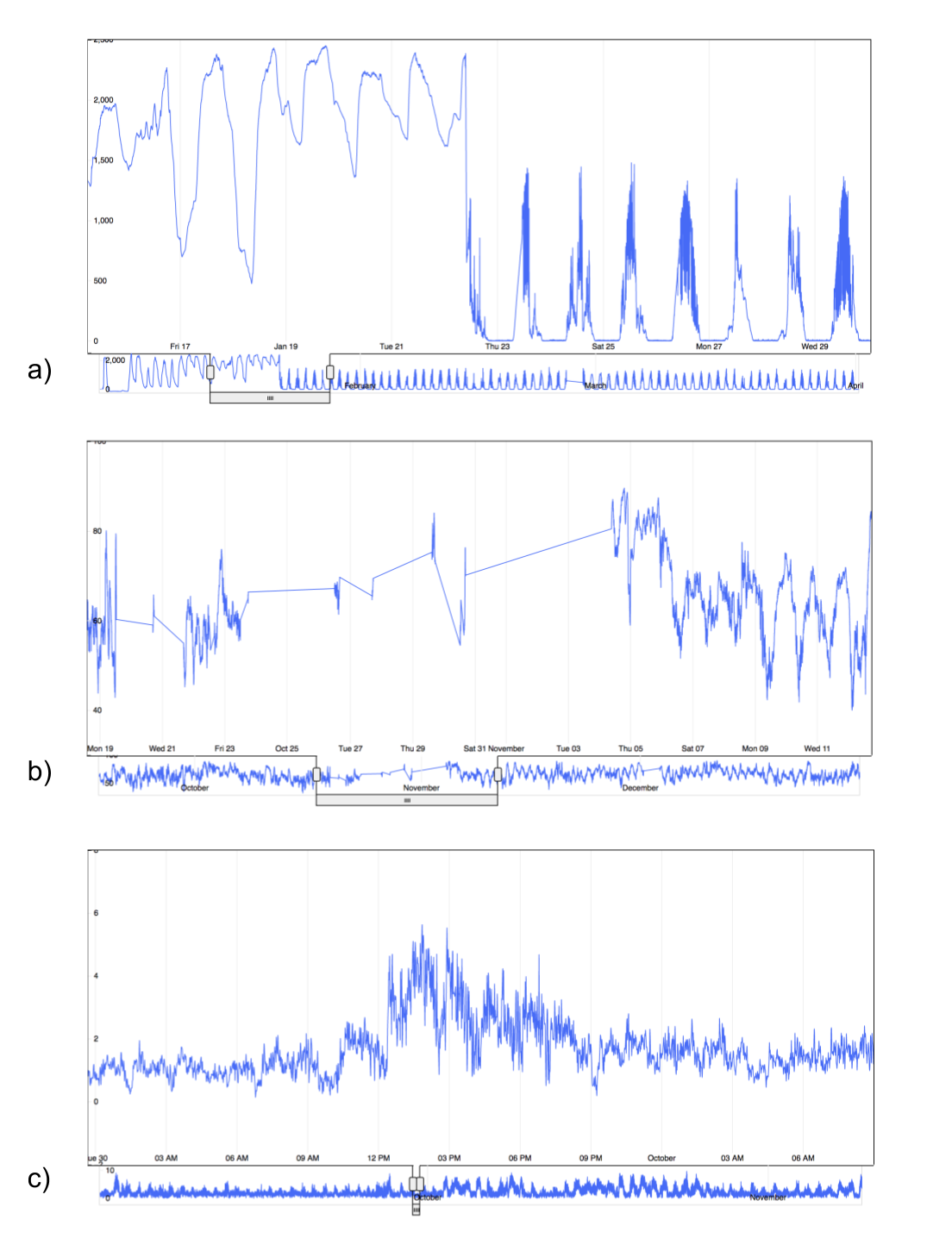}
 \caption{Our time series inspection tool, with a context bar and focus selection. \emph{a)} The heading $I_D$ and $I_G$ values for \emph{Moufia} are governed by an unusual stochastic process. \emph{b)} Straight lines facilitate the visual identification of time ranges to exclude, here in the \emph{Moufia RH} variable. \emph{c)} The 3 isolated missing timestamps on \texttt{2014-09-30} in \emph{Possession} are visually imperceptible.}
 \label{fig:tsviz}
\end{center}
\end{figure}

The data files were completed with missing timestamps, and missing values were linearly interpolated.
Whenever the interpolated range is too large, undesirable artifacts result from this procedure (straight lines, or pure sine curves for \emph{UnitX} and \emph{UnitY}). The associated time frames should be excluded from further analysis for the respective site. However, when a single time slot is missing, the effect of interpolation on the time series shape is imperceptible. We customized a time series visualization tool (\cite{tsline12}, see Figure \ref{fig:tsviz}), and identified the relevant exclusion frames visually. They are reported in Table \ref{tab:exclusion}.

\begin{table}
\begin{center}
\begin{tabular}{|l|l|}
  \hline
  Site & Excluded Slots \\
  \hline
  Moufia & 2014-01-01 to 2014-01-23 \\
    & 2014-02-25 to 2014-02-28 \\
    & 2015-10-19 to 2015-11-05 \\
    & 2015-12-01 to 2015-12-04 \\
  \hline
  Saint André & 2014-01-01 to 2014-03-29 \\
  \hline
\end{tabular}
\end{center}
\caption{Frames excluded from analysis for each site. For simplicity and safety, the slots were rounded to the closest enclosing day. Absent sites have no excluded frame.}
\label{tab:exclusion}
\end{table}

Almost all excluded frames emerge from missing values or timestamps, except the heading values in the \emph{Moufia} data. The stochastic process governing the latter is visibly different from \emph{usual} irradiation values (see Figure \ref{fig:tsviz}a), we thus added the associated timestamps to the excluded frames.
Continuing with the visual exploration, we observed that to the contrary of all other variables, the irradiance series exhibit a strong prior at night times. In practice, nocturnal irradiance values are close to zero, and sensor imprecision in this range leads to inconsistant $k_b$ values.
We alleviate this issue by setting:

\begin{equation}
  k_b = 0.5 + e \text{, with } e \sim \mathcal N(0, \sigma^2)
\end{equation}

at night-time, by doing so controlling the variance of noisy night values, hence limiting their influence in further processing. The nocturnal timestamps are determined using publicly available sun rise and set times for \emph{La Réunion} island \cite{sunrise}.

\subsection{Variable Selection} \label{sec:selection}
As stated by the end of Section \ref{sec:data}, hourly predictions in the context of time series with 1-minute time steps boils down to predicting $x_{\delta, t+59}$ with the knowledge of all time series $x_d$ up to timestamp $t-1$.
Even before considering a specific prediction function structure, we face some important design choices:

\begin{itemize}
\item use only the $D$ time series values at time $t-1$ as input,
\item if adding past values, define a strategy to select a reasonable input vector size.
\end{itemize}

Indeed, in general the input dimensionality of a prediction function has significant influence on training time and requirements in terms of training set sizes. In the case of the MLP model, the training time is quadratic w.r.t. to the number of input dimensions \cite{bruneau12a}, and linear at best if a \emph{Stochastic Gradient Descent} (SGD) method is used \cite{zhang04}.
In other words, the input vector size has to be sufficiently large so as to yield satisfactory performance, but remain as small as possible to avoid prohibitively large training time, as well as excessive consumption of central memory.

As already mentionned in Section \ref{sec:related}, ARIMA models have an established methodology for variable selection. We propose to perform a set of univariate ARIMA analyses (i.e. each analysis uses a single time series). We use the stepwise selection method proposed by \cite{hyndman07}, and implemented in the \emph{forecast} package in R, to pre-select relevant past values for each time series.
In principle, the resulting selection improper in the context of hourly predictions, as it is optimized w.r.t. 1-step-ahead predictions.
However, we hypothesize that this univariate selection method provides general purpose machine learning regression models with a useful input, in view of predicting ${k_b}_{t+59}$. Implicitly, the account of covariates between meteorological variables is delegated to the general-propose model. We perform this analysis on the data for year 2014 taken from the \emph{Possession} site.

In practice, estimating ARIMA models with orders larger than 30 (i.e. the value of $p$ or $q$ in Equations \eqref{eq:arp} and \eqref{eq:maq}) is computationally demanding. However, meteorological variables are \emph{a priori} expected to exhibit a 24-hour seasonality. This would require estimating ARIMA models with orders at least as high as 1440, which is impossible in practice.

To alleviate this issue, we subsampled the hourly values in the provided data (i.e. those values associated with timestamps with both \emph{minutes} and \emph{seconds} fields at 0). We also estimated ARIMA models on these subsampled time series, that optimize the loss when forecasting $x_{t+59}$ with the knowledge of $x_{t-1}$, $x_{t-61}$, etc...

\begin{figure}[H]
\begin{center}
 \includegraphics[width=0.9\textwidth, keepaspectratio]{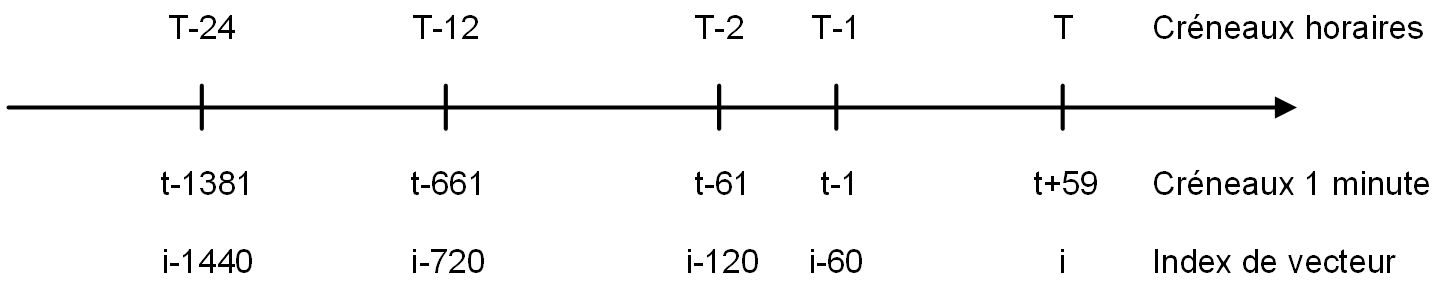}
 \caption{Conversion between hourly and 1-minute timestamps. As programmatic contexts often use the predicting horizon as reference time, the conversion to array indices is also reported.}
 \label{fig:conversion}
\end{center}
\end{figure}

For convenience, we define the time index $T$ that refers to hourly periodicity, when $t$ refers to 1-minute periods. The conversion between these two indices is illustrated in Figure \ref{fig:conversion}. We highlight that with this indexing system, $x_{t-1}$ designates the same value as $x_{T-1}$, and that the hourly predicting horizon is now equivalently designated by $x_{t+59}$ and $x_T$.

Following the automatic estimation procedure defined in \cite{hyndman07} on both hourly and 1-minute series, we diagnose their correctness using auto-correlation plots. From the starting point obtained by automatic means, terms were added until the residuals and auto-correlation plots were satisfactory. Assuming $\text{ARIMA}(p,d,q)$ has been estimated, we inferred that $\max(p,q)$ past terms (i.e. $x_{t-1} \dots x_{t-\max(p,q)}$) are selected in order to feed the input vector for machine learning. For computational reasons, $\max(p,q)$ was limited to 20.

The hourly and seasonal terms are selected in a similar way. The resulting selection is summarized in Table \ref{tab:selection}. We note that all selections are fairly parsimonious, except for the \emph{Patm} variable. Chances are the optimal order would be even higher than 20 then. The resulting input vector, that aggregates all slots selected for all variables, has 70 dimensions.

\begin{table}
\begin{center}
\begin{tabular}{|l|l|}
  \hline
  Variable name & Selected slots \\
  \hline
  $k_b$ & $t-1, t-2, T-2, T-3, T-24$  \\
  \emph{Patm} & $t-1, \dots, t-20, T-2, \dots, T-6, T-12, T-24$  \\
  \emph{RH} & $t-1, \dots, t-5, T-24$ \\
  \emph{Text} & $t-1, t-2, t-3, T-12, T-24$ \\
  \emph{WS\_Mean} & $t-1, t-2, t-3, T-2, T-3, T-12, T-24$ \\
  \emph{UnitX} & $t-1, \dots, t-5, T-2, T-3, T-4, T-12, T-24$ \\
  \emph{UnitY} & $t-1, \dots, t-6, T-2, T-3, T-12, T-24$ \\
  \hline
\end{tabular}
\end{center}
\caption{The time slots selected by ARIMA for each variable. NB: as $T-1$ and $t-1$ are equivalent, $t-1$ only is indicated when relevant.}
\label{tab:selection}
\end{table}

All estimated ARIMA models used a differentiated version of the time series. This means that the original data was unlikely to emerge from a stationary process (i.e. constant mean and variance). A straightforward mean to establish weak stationarity, when using the selected variables as input of the models presented in next section, is to normalize the series data with monthly-hourly means, as introduced in \cite{bruneau12a}:

\begin{align}
& x_t^\text{norm} = \frac { x_t - \text{mean}(\textbf x_{m^*,h^*}) } { \text{standard deviation}(\textbf x_{m^*,h^*}) } \label{eq:months-hours}\\
& \text{with }m^* = \text{month}(t), h^* = \text{hour}(t) \nonumber \\
& \text{and }\textbf x_{m, h} = \{ x_t | \text{month}(t) = m \wedge \text{hour}(t) = h \} \nonumber
\end{align}

\section{Experiments} \label{sec:experiments}

\subsection{Protocol}

As introduced earlier in Section \ref{sec:selection}, and summarized in Figure \ref{fig:conversion}, for a given present time $T-1$ this experimental part is focused on the prediction of $k_b$ for the timestamp $T$. We will test the following hypotheses:

\begin{itemize}
\item whether the variable selection among time slots $t^*<t-1$ performs better than using only the instant values at $t-1$ as input,
\item whether a model learned for a given geographical site performs better on the test set for its site than one learned using data from another site.
\end{itemize}

In the remainder, we will designate the input made of the 7 time series values at $t-1$ as the \emph{instant} input vector, and the input resulting from the variable selection procedure described in Section \ref{sec:selection} as the \emph{arima} vector.
To test the hypotheses above, for each site we will use the data for year 2014 as a training set, and the data for year 2015 as test set. In these sets, we exclude the items associated to prediction targets falling on nocturnal time slots, as solar irradiation has no value then. We will use the two following general-purpose models and respective training algorithms:

\begin{itemize}
\item \emph{Xgboost}: this learning technique uses Gradient Boosting \cite{friedman01} and Generalized Boosted Models \cite{ridgeway07}.
We implemented it using the \emph{Xgboost} R package \cite{chen16}. This algorithm involves several hyper-parameters, such as the learning rate $\eta$, the maximum depth of a tree, and $\gamma$, the minimum loss reduction required to split a node. These parameters have been tuned in R using the \emph{caret} package \cite{kuhn15} using 3-fold cross validation. Predictions were then obtained from the best model estimated on the complete training set. Experimentally, it was found that Xgboost performs much better using unnormalized variables, so neither basic variable standardization, nor monthly-hourly sets described by Equation \eqref{eq:months-hours} were used in Xgboost experiments.
\item \emph{MLP}: this neural network architecture uses a single hidden layer. We implemented it using the \emph{mxnet} R package \cite{chen15}. The optimal model complexities (i.e. number of neurons in the hidden layer) for \emph{instant} and \emph{arima} vectors, respectively, were determined using 10-fold cross-validation on the \emph{Possession} site data. The candidate MLP hidden layer sizes were taken from the range $\{5, \dots, 1000\}$. Values eventually selected are rather low (10 and 30 for \emph{instant} and \emph{arima} vectors, respectively). The MLP complexities roughly correlate to the input vector size. Models were optimized for each site using the appropriate complexity, again with 10-fold cross-validation. As neural networks are very sensitive to variable normalization, the time series data used to build the \emph{instant} and \emph{arima} vectors were normalized according to monthly-hourly sets, as described in Equation \eqref{eq:months-hours}. Predictions were then obtained from the model ensemble made by the 5 best models in terms of validation error.
\end{itemize}

\subsection{Results}
In Table \ref{tab:results}, we report the test \emph{Root-Mean-Square Error} (RMSE) and \emph{Mean Absolute Error} (MAE) obtained by models trained for single sites. RMSE is reported as it is the metric that is generally optimized by regression models. MAE is also reported as a support to interpretation, as it is the average error to be expected for a single prediction.

\begin{table}
\begin{center}
\begin{tabular}{ cll|l|l|l|l|l| }
\cline{4-8}
 & & & Moufia & Possession & Saint & Saint & Saint \\
 & & & &  & André & Leu & Pierre \\
\hline
\multicolumn{1} { |c| }{\multirow{4}{*}{instant}} &
  \multirow{2}{*}{RMSE} & \multicolumn{1}{ |l| }{Xgboost} & 0.268 & 0.266 & 0.271 & 0.278 & 0.264 \\
\multicolumn{1} { |c| }{}  & & \multicolumn{1}{ |l| }{MLP} & 0.318 & 0.332 & 0.316 & 0.384 & 0.355 \\
  \cline{2-8}
\multicolumn{1} { |c| }{} & \multirow{2}{*}{MAE} & \multicolumn{1}{ |l| }{Xgboost} & 0.208 & 0.211 & 0.217 & 0.222 & 0.201 \\
\multicolumn{1} { |c| }{} & & \multicolumn{1}{ |l| }{MLP} & 0.204 & 0.246 & 0.225 & 0.288 & 0.263 \\
\hline
\multicolumn{1} { |c| }{\multirow{4}{*}{arima}} &
  \multirow{2}{*}{RMSE} & \multicolumn{1}{ |l| }{Xgboost} & \textbf{0.249} & \textbf{0.245} & \textbf{0.257} & \textbf{0.255} & \textbf{0.241} \\
\multicolumn{1} { |c| }{}  & & \multicolumn{1}{ |l| }{MLP} & 0.283 & 0.283 & 0.332 & 0.315 & 0.286 \\
  \cline{2-8}
\multicolumn{1} { |c| }{} & \multirow{2}{*}{MAE} & \multicolumn{1}{ |l| }{Xgboost} & \textbf{0.193} & \textbf{0.195} & \textbf{0.208} & \textbf{0.203} & \textbf{0.183} \\
\multicolumn{1} { |c| }{} & & \multicolumn{1}{ |l| }{MLP} & 0.202 & 0.214 & 0.255 & 0.243 & 0.205 \\
\hline
\end{tabular}
\end{center}
\caption{Errors obtained on test sets for \emph{instant} and \emph{arima} vectors. The lowest RMSE and MAE for each site are bold-faced.}
\label{tab:results}
\end{table}

Using the \emph{arima} vector yields MLP and Xgboost models with significantly better RMSE, except for the MLP of the \emph{Saint André} site. However the performance gain is moderate: as $k_b$ scales in $[0,1]$, at best a single prediction is expected to have 18.3\% error using Xgboost models (20.2\% with MLP models). Also, the improvement brought by using the \emph{arima} vector is limited with regard to the influence of the regression function. For example, Xgboost that uses the \emph{instant} vector is always better than the MLP model that uses the \emph{arima} vector.

In the remainder, we retain the models specific to each site trained with the \emph{arima} vector. Prior to applying the models on test data for all sites, and thus estimating their generalization ability, Table \ref{tab:correlation} reports the average correlation over time, and between sites, for the 7 meteorological variables considered in this study. Quite naturally, the time series that encode wind information (\emph{WS\_Mean}, \emph{UnitX} and \emph{UnitY}) have very weak correlations, even close to 0 for the direction variables. \emph{Text} and \emph{Patm} exhibit the strongest correlations, which is also quite natural, as external temperature and atmospheric pressure are expected to be quite homogeneous over a territory as small as \emph{La Réunion} island. However, the high variance associated to \emph{Patm} moderates this homogeneity, rather suggesting a partial homogeneity, e.g. two homogeneous groups from the perspective of \emph{Patm}.

Overall, Table \ref{tab:correlation} reflects the fact that the sites (and their respective data) cannot be aggregated as a single site \emph{a priori}. This conjecture is reinforced by the fact that the target variable, $k_b$, is rather weakly correlated across sites.

Table \ref{tab:cross} displays the RMSE values obtained when evaluating models specific to a site on data from other sites. The diagonal in this table is then made of the RMSE values displayed in Table \ref{tab:results}. Reading lines in this table shows how a given model performs on all sites. The most striking observation is that Xgboost models perform consistently worse than on their genuine site, when MLP models perform well on all sites. Variable normalization is a possible explanation for this phenomenon: monthly-hourly sets may set all variables to a common scale, thus encoding general climate properties. On the other hand, Xgboost splits refer to the original data scale, which is likely to differ sensibly between sites.
Also interesting is the fact that the best performance of a MLP is sometimes not obtained on its genuine site (e.g. the MLP for \emph{Sain Leu} get its best score on \emph{Saint Pierre}), and the best performing model for a site is not always the model trained on this site (e.g. the MLP for \emph{Possession} is the best MLP for 4 sites). A possible explanation is that the variable selection procedure and MLP complexity validation were carried out on the data for \emph{Possession}.

\begin{table}
\begin{center}
\begin{tabular}{|l|l|l|l}
  \hline
  $k_b$ & \emph{Patm} & \emph{RH} & \multicolumn{1}{ |l| }{\emph{Text}} \\
  \hline
  0.44 $\pm$ 0.09 & 0.67 $\pm$ 0.41 & 0.57 $\pm$ 0.08 & \multicolumn{1}{ |l| }{0.89 $\pm$ 0.02} \\
  \hline
  \hline
  \emph{WS\_Mean} & \emph{UnitX} & \emph{UnitY} & \\
  \cline{1-3}
  0.23 $\pm$ 0.13 & -0.04 $\pm$ 0.36 & -0.01 $\pm$ 0.33 & \\
  \cline{1-3}
\end{tabular}
\end{center}
\caption{Average correlation between sites for the studied time series. Standard deviations are reported as an indication of the values distribution.}
\label{tab:correlation}
\end{table}

\begin{table}
\begin{center}
\begin{tabular}{ cl|l|l|l|l|l| }
\hline
\multicolumn{2} { |c| }{\multirow{2}{*}{\backslashbox{Train}{Test}}} & Moufia & Possession & Saint & Saint & Saint \\
\multicolumn{2} { |c| }{} & & & André & Leu & Pierre \\
\hline
\multicolumn{1} { |c| }{\multirow{2}{*}{Moufia}} &
  \multicolumn{1}{ |l| }{Xgboost} & 0.249 & 0.392 & 0.353 & 0.340 & 0.365 \\
\multicolumn{1} { |c| }{}  & \multicolumn{1}{ |l| }{MLP} & 0.283 & 0.297 & 0.291 & 0.317 & 0.288 \\
\hline
\multicolumn{1} { |c| }{\multirow{2}{*}{Possession}} &
  \multicolumn{1}{ |l| }{Xgboost} & 0.307 & 0.245 & 0.306 & 0.291 & 0.274 \\
\multicolumn{1} { |c| }{}  & \multicolumn{1}{ |l| }{MLP} & \textbf{0.272} & 0.283 & \textbf{0.276} & \textbf{0.303} & 0.290 \\
\hline
\multicolumn{1} { |c| }{Saint} &
  \multicolumn{1}{ |l| }{Xgboost} & 0.318 & 0.439 & 0.257 & 0.302 & 0.398 \\
\multicolumn{1} { |c| }{André}  & \multicolumn{1}{ |l| }{MLP} & 0.330 & 0.351 & 0.332 & 0.353 & 0.320 \\
\hline
\multicolumn{1} { |c| }{Saint} &
  \multicolumn{1}{ |l| }{Xgboost} & 0.278 & 0.347 & 0.295 & 0.255 & 0.339 \\
\multicolumn{1} { |c| }{Leu}  & \multicolumn{1}{ |l| }{MLP} & 0.307 & 0.315 & 0.312 & 0.315 & 0.289 \\
\hline
\multicolumn{1} { |c| }{Saint} &
  \multicolumn{1}{ |l| }{Xgboost} & 0.330 & 0.314 & 0.306 & 0.310 & 0.241 \\
\multicolumn{1} { |c| }{Pierre}  & \multicolumn{1}{ |l| }{MLP} & 0.306 & 0.313 & 0.303 & 0.317 & 0.286 \\
\hline
\end{tabular}
\end{center}
\caption{RMSE for between-site predictions using the \emph{arima} vector. Sites in rows indicate the data set used to train the respective model, sites in columns indicate the test set used for prediction. Situations where the best model for a test set is trained on another site are bold-faced.}
\label{tab:cross}
\end{table}


\section{Conclusion}

We adressed \emph{Défi EGC 2018} under the perspective of short-term prediction of solar irradiance, namely the prediction of $k_b$ one hour ahead.
We presented a pre-processing pipeline, a variable selection procedure based upon the ARIMA model, and experimental results obtained with MLP and Xgboost models. We observed that Xgboost performs better for the task at hand. Our selection procedure improves the results for both models, but the improvement can be seen as marginal to that of using Xgboost instead of MLP.

A short study of correlations between sites shows that data sets for the 5 sites cannot be trivially merged prior to learning prediction models. After training models on data from their respective sites, we also monitored how these site-specific models performed on test sets from other sites. Using monthly-hourly normalization brings an advantage to MLP models then. However the collected evidence suggests that the variable selection procedure should be applied to each site separately in order to bring optimal performance.

The general approach we adopted in this paper takes little account of the sequential nature of the data. A promising way for improvement would be to use a model that accounts for this structure, for example by adapting convolutional neural networks \cite{krizhevsky12}, or recurrent neural networks \cite{williams89} to the short term prediction task described above. These models have indeed been successfully used in the context of sequential problems (e.g. textual sentiment analysis \cite{severyn15}, machine translation \cite{liu14}).

\bibliographystyle{plain}
\bibliography{refs}

\end{document}